\documentclass[letterpaper, 10 pt, journal, twoside]{IEEEtran}

\usepackage{amsmath,amsfonts}
\usepackage{algorithm}
\usepackage{array}
\usepackage{textcomp}
\usepackage{url}
\usepackage{verbatim}
\usepackage{graphicx}
\usepackage{xcolor}
\usepackage{cite}
\usepackage{tikz}
\usepackage{hyperref}
\usetikzlibrary{arrows.meta,calc,decorations.pathreplacing,patterns.meta}

\usepackage{algpseudocode}
\usepackage{booktabs}
\usepackage{subfigure}
\usepackage{tabularx}
\usepackage[british]{babel}
\usepackage[autostyle]{csquotes}

\begin{document}
\bstctlcite{BSTcontrol}

\title{C-ARC: Continuous-Adaptive Range Clustering for Non-Repetitive LiDAR Sensors}

\author{
    Nick B. Schroeder$^{*}$,
    Jonathan Lichtenfeld$^{*}$,
    and Oskar von Stryk%
    \thanks{© 2026 IEEE. Personal use of this material is permitted.
    Permission from IEEE must be obtained for all other uses,
    in any current or future media, including
    reprinting/republishing this material for advertising
    or promotional purposes, creating new collective works,
    for resale or redistribution to servers or lists, or
    reuse of any copyrighted component of this work in other works.}%
    \thanks{Manuscript received: May 29, 2026; Revised August 20, 2026; Accepted September 15, 2026.}%
    \thanks{Publication available at IEEE Xplore: \href{https://doi.org/10.1109/LRA.2026.3742307}{doi:10.1109/LRA.2026.3742307}.}%
    \thanks{This article was recommended for publication by Associate Editor M. Mattamala and Editor A. Kim upon evaluation of the reviewers' comments.
    This work was supported in parts by the LOEWE initiative (Hesse, Germany) within the emergenCITY center [LOEWE/1/12/519/03/05.001(0016)/72] and by the Federal Ministry of Research, Technology and Space (BMFTR) within the projects DRZ (grant no.~13N16475) and KIARA PRO (grant no.~13N17765).}%
    \thanks{$^{*}$Equal contribution.}%
    \thanks{The authors are with the SIM Group, Technical University of Darmstadt, Germany. {\tt\small \{lichtenfeld,stryk\}@sim.tu-darmstadt.de}}%
    \thanks{Code will be released at \protect\url{https://github.com/tu-darmstadt-ros-pkg/C-ARC}.}%
}

\markboth{IEEE Robotics and Automation Letters. Preprint Version. Accepted September, 2026}%
{Schroeder \MakeLowercase{\textit{et al.}}: C-ARC: Continuous-Adaptive Range Clustering for Non-Repetitive LiDAR Sensors}


\maketitle



\begin{abstract}
	Real-time LiDAR clustering identifies structures in point clouds, which is an essential prerequisite for many mobile robotics algorithms. 
    Current methods are mostly developed for repetitive mechanical LiDAR sensors.
    Recently, the use of non-repetitive LiDAR sensors is strongly increasing due to their low cost and form factor.
    Such non-repetitive Risley prism-based sensors violate two key assumptions of repetitive mechanical sensors: structured scan lines and well-defined frame boundaries.
    Their Rhodonea-curve trajectories produce non-uniform point distributions, and the absence of a rotation cycle renders conventional scan line indexing inapplicable.
    To meet such new requirements, we developed C-ARC, a Continuous-Adaptive Range Clustering framework that maintains a persistent dual-graph over a sliding window, decoupling high-frequency point insertion from on-demand cluster retrieval.
    This is crucial for key functionalities like SLAM or tracking.
	An adaptive range-grid sizing step selects a single angular cell size matching the sensor FOV aspect so that mean cell multiplicity stays within a mild overload threshold, without prior knowledge of the scanning pattern.
    Evaluations show competitive segmentation quality on the Livox Mid-360 and a spinning LiDAR, and real-time operation on the Mid-360; on the Livox Avia, unbounded cell occupancy remains the main limit on its concentrated pattern.
	The same sizing transfers across sensors and also improves clustering quality for existing grid-based methods on non-repetitive data.
\end{abstract}



\section{INTRODUCTION}
\IEEEPARstart{L}{iDAR} is a fundamental perception technology in autonomous systems, providing dense 3D measurements of the environment at high frequency.
Foundational tasks such as segmentation, object detection, and clustering must be processed in real time to support downstream applications like SLAM~\cite{dube2017SegMatchSegmentBased, lichtenfeld2024EfficientDynamicLiDAR}, as latency directly affects planning and control~\cite{li2020LidarAutonomousDriving}.
\begin{figure}[!t]
	\centerline{
		\subfigure[]{%
			\includegraphics[width=0.33\columnwidth,trim=0 0 0 52,clip]{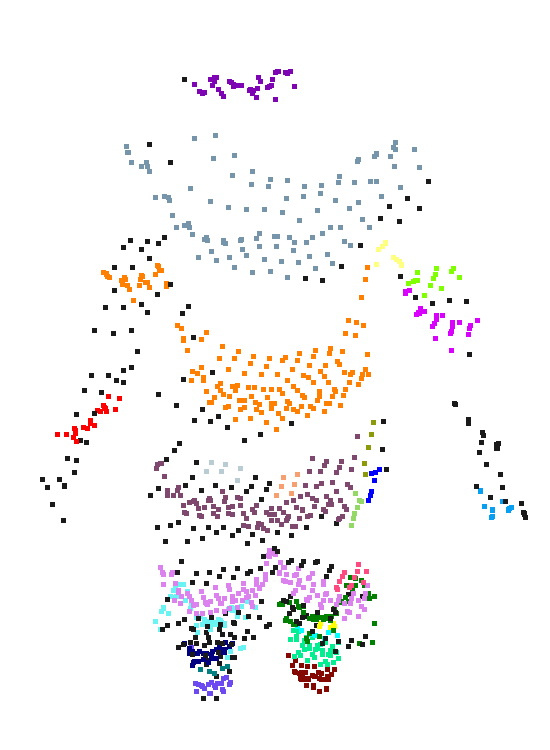}\label{fig:healing:1}
		}\hfil
		\subfigure[]{%
			\includegraphics[width=0.33\columnwidth,trim=0 0 0 52,clip]{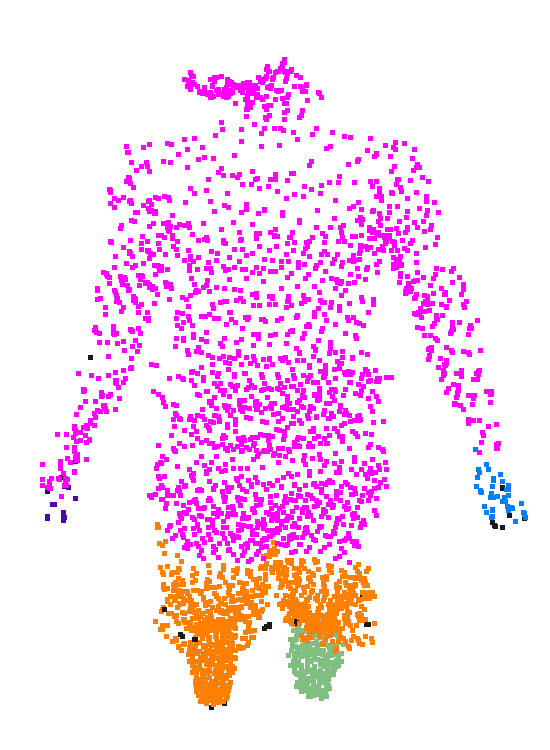}\label{fig:healing:2}
		}\hfil
		\subfigure[]{%
			\includegraphics[width=0.33\columnwidth,trim=0 0 0 52,clip]{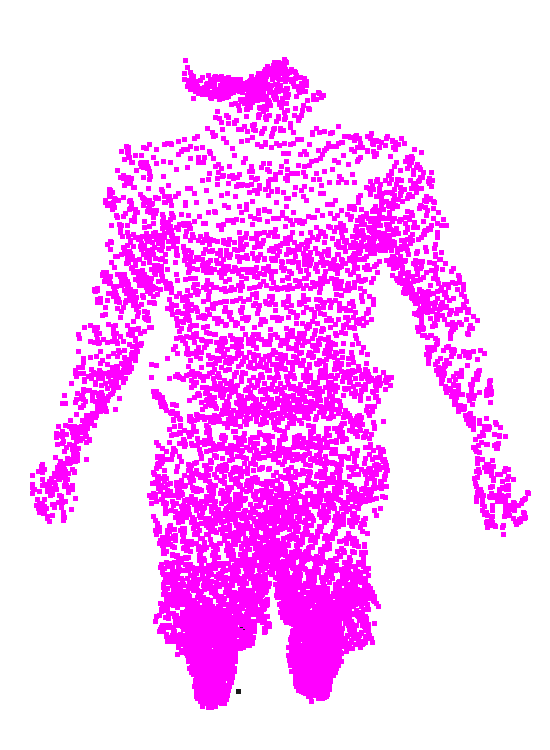}\label{fig:healing:3}
		}
	}
	\caption{At system start, a single object may be represented by many disconnected clusters~(a).
		As the buffer fills, fragmented clusters are gradually merged into larger, more stable connected components~(b), eventually forming fully connected high-level components~(c).
	}\label{fig:healing}
\end{figure}

The dominant paradigm for real-time LiDAR clustering has been built around mechanical rotating sensors, which produce structured, repetitive scan lines at a fixed angular resolution.
Algorithms leverage this known scanning structure to achieve constant-time nearest neighbour search (NNS) via range image projection, defining discrete processing frames around the sensor's rotation period.
The resulting methods are highly efficient, but their performance is fundamentally contingent on the structured nature of repetitive scan data.

Non-repetitive LiDAR sensors, such as the Livox series, offer an alternative mechanism.
A pair of counter-rotating prisms steers the laser beams, producing a Rhodonea-curve, a petal-shaped scanning pattern whose density varies across the Field of View (FOV)~\cite{brazeal2021RigorousObservationModel}.
Because the ratio of the two prism rotation speeds is irrational, the pattern does not repeat, yielding progressively increasing spatial coverage over time.

These sensors offer competitive FOV at substantially lower cost and weight than traditional spinning scanners and are increasingly explored in resource-constrained mobile robotics platforms
~\cite{
	ren2025SafetyassuredHighspeedNavigation}.

Despite this adoption, the non-uniform point distribution of Risley prism-based sensors poses a fundamental challenge for real-time clustering.
Two properties of the scanning pattern break the assumptions of standard methods.
First, the absence of discrete scan lines makes \textit{Projection by ID} (PBID) indexing~\cite{wu2021DetailedAnalysisGenerating} inapplicable:
PBID maps each scan line of a repetitive sensor to a dense range image row without overlapping points, but without a fixed scan structure, the resulting projection is both sparse and non-uniform.
Second, without a mechanical rotation cycle, there is no intuitive frame boundary. 
The choice of the accumulation window therefore becomes critical. 
Longer windows improve point density and cluster connectivity (Fig.~\ref{fig:healing}), but increase latency and motion distortion. 
Shorter windows reduce these effects, yet produce projections that are too sparse for reliable connectivity. 
This creates an inherent trade-off between latency and point density.

Existing continuous clustering methods~\cite{najdataei2017LiscoContinuousApproach, najdataei2022PiLiscoParallelIncremental, reich2024LowLatencyInstance} address the latency problem by maintaining a sliding window over the point stream but retain PBID-style indexing and are therefore restricted to repetitive sensors.
To the authors' best knowledge, no existing framework combines continuous, point-wise clustering with a data structure that accommodates the irregular density characteristics of Risley prism-based sensors.

To address this gap, this paper makes the following contributions:

\begin{enumerate}
	\item \textbf{Adaptive Range Grid Sizing:} A sensor-agnostic procedure that chooses a single FOV-aspect angular cell size so mean cell multiplicity stays below a load threshold $T_M$, without prior knowledge of the scanning pattern.
	      The same procedure can initialize any grid-based clustering algorithm.

	\item \textbf{C-ARC Framework:} A continuous point-wise clustering framework that maintains a persistent dual-graph over a sliding window $T_B$.
	      Lazy deletion defers structural validation to retrieval, decoupling high-frequency insertion from lower-frequency publication and removing frame-accumulation latency.
\end{enumerate}


\section{RELATED WORK}
\label{sec:related_work}
\noindent We group prior work into three areas: point cloud segmentation, exploitation of non-repetitive scan patterns, and discrete versus continuous processing paradigms.

\subsection{LiDAR Point Cloud Segmentation}
\noindent Clustering efficiency is bounded by the underlying NNS, so we group methods by neighbourhood representation: tree-based, grid-based, and density-based.
Tree-based structures such as \textit{k}-d trees and incremental variants like ikd-Tree~\cite{cai2021IkdTreeIncrementalKD} provide logarithmic average-case complexity but require continuous insertions and structural maintenance in streaming LiDAR applications.
Systems such as FAST-LIO2~\cite{xu2022FASTLIO2FastDirect} demonstrate real-time capability with per-point insertion times of a few microseconds.
However, at high input rates (e.g.\,$~$200\,kHz), this cost approaches the available processing budget, leaving limited headroom for clustering.
Tree-based methods also exhibit latency spikes due to rebalancing and irregular memory access patterns~\cite{xu2022FASTLIO2FastDirect}, which is problematic for continuous clustering frameworks requiring predictable per-point update costs.

Grid-based representations instead provide constant-time $O(1)$ neighbourhood access through spatial discretization.
Angular $H \times W$ layouts are common in LiDAR segmentation.
We distinguish two uses of the same spherical indexing:
a \emph{range image} stores at most one return per cell (later points overwrite earlier ones)~\cite{bogoslavskyi2016FastRangeImagebased, moosmann2009Segmentation3DLidar},
whereas a \emph{range grid} accumulates multiple points per cell in a bucket (Sec.~III).
Range image methods typically reduce 3D clustering to 2D connected-component labelling (CCL)~\cite{he2017ConnectedcomponentLabelingProblem}; 
a range grid instead also supports neighbourhood search over the retained points in each cell.
Occupied Cartesian or polar voxels, e.g.\ CVC~\cite{parkCurvedVoxelClusteringAccurate2019b}, cluster cells rather than a 2D range image; purely 3D methods such as RBNN~\cite{klasingClusteringMethodEfficient2008} cluster in Euclidean space without projection.

Density-based methods such as DBSCAN~\cite{ester1996dbscan} and streaming variants including DenStream~\cite{cao2006DensityBasedClusteringEvolving} are well suited to unstructured point clouds because they do not depend on scan-line structure.
However, they require repeated neighbourhood queries and cluster expansion operations, typically supported by dynamic nearest neighbour structures, resulting in substantial overhead in high-frequency LiDAR streams.
Their density estimates can also be sensitive to the irregular and evolving sampling patterns of non-repetitive sensors.

Most existing projection-based LiDAR segmentation methods exploit the structured scan patterns of repetitive mechanical sensors.
Using PBID indexing~\cite{wu2021DetailedAnalysisGenerating}, scan rings can be mapped directly to dense range images without overlap.
Methods such as~\cite{moosmann2009Segmentation3DLidar,bogoslavskyi2016FastRangeImagebased} perform efficient breadth-first search (BFS) or CCL on range images, while later works improve throughput through two-pass cluster merging~\cite{zermas2017FastSegmentation3D,yang2020TwoLayerGraphClusteringRealTime} or scan-ring-based geometric feature extraction~\cite{li2025ComplexGeodesicCurvature}.
Although highly efficient, these approaches remain strongly dependent on structured scanning geometries and are therefore difficult to generalize to non-repetitive sensors.

\subsection{Non-Repetitive Scan Exploitation}
\noindent As described in Sec.~I, non-repetitive LiDAR sensors such as the Livox series produce a deterministic Rhodonea trajectory whose coverage grows over time.
Although structured, these sensors lack discrete scan lines, making conventional scan line–based indexing unsuitable for many online clustering methods.

In perception applications, such sensors have shown promise for SLAM and object detection due to their wide FOV and increasing sampling density~\cite{volpin2023360degFOVScanning, xie2024NonRepetitivePromisingLiDAR}.
Recent works~\cite{aijazi2024NonRepetitiveScanningLiDAR, aijazi2025BuildingFastDynamic} exploit the centre of the Rhodonea pattern as a reference for point cloud registration and trajectory planning, indicating that pattern-aware methods can reduce runtime complexity.

Existing research primarily models these scanning patterns through prism kinematics~\cite{li2025AdvancesRotatingRisley} or empirical observation models~\cite{brazeal2021RigorousObservationModel}.
However, real-time kinematic modelling and online recalibration introduce additional computational overhead, limiting suitability for latency-critical applications.

Despite these advances, little work has investigated how the non-uniform sampling behaviour of such sensors affects connectivity-based clustering.

\subsection{Continuous vs. Discrete Paradigm}
\noindent Standard online clustering methods~\cite{bogoslavskyi2016FastRangeImagebased, 
moosmann2009Segmentation3DLidar, 
zermas2017FastSegmentation3D, 
yang2020TwoLayerGraphClusteringRealTime, 
zhang2022RealTimeFastChannel, 
oh2022TRAVELTraversableGround} 
operate on discrete frames, introducing an inherent latency of
$\Delta t_\mathrm{l} = \Delta t_\mathrm{F} + \Delta t_\mathrm{P}$. Even with negligible processing time $t_\mathrm{P}$, latency remains bounded by the frame accumulation period $\Delta t_\mathrm{F}$, typically 50--100~ms. Partitioning frames into sub-scans can reduce latency~\cite{han2020StreamingObjectDetection} 
but may fragment clusters at sub-scan boundaries~\cite{reich2024LowLatencyInstance}.

Continuous clustering mitigates this limitation by maintaining a sliding window over the incoming point stream rather than processing discrete frames~\cite{al-khamees2021SurveyClusteringTechniques}.
Early approaches such as \textit{Lisco}~\cite{najdataei2017LiscoContinuousApproach} still accumulated full sensor sweeps before publishing clusters, whereas later methods~\cite{najdataei2022PiLiscoParallelIncremental, reich2024LowLatencyInstance} achieved point-wise updates using overwritable or expanding range images.
Reich and Maehlisch~\cite{reich2024LowLatencyInstance} publish an instance once the spinning sweep has covered it, rather than waiting for a full revolution or a fractional reporting period as in Pi-Lisco~\cite{najdataei2022PiLiscoParallelIncremental}.

However, these continuous methods still update clusters eagerly on insertion and retain PBID-style layouts, so they leave the non-repetitive clustering gap above open.
Non-repetitive patterns also offer no sweep-coverage event of the kind Reich uses for release timing.
Tab.~\ref{tab:method_positioning} summarizes the resulting design space.

\begin{table}[!ht]
    \caption{Positioning of related clustering methods.}
    \label{tab:method_positioning}
    \centering
    \footnotesize
    \setlength{\tabcolsep}{2pt}
    \begin{tabularx}{\columnwidth}{@{}l >{\centering\arraybackslash}X c c >{\centering\arraybackslash}X@{}}
        \toprule
        \textbf{Method} & \textbf{Projection} & \textbf{Layout}\textsuperscript{*} & \textbf{Cont.}\textsuperscript{*} & \textbf{Publish}\textsuperscript{*} \\
        \midrule
        Bogoslavskyi~\cite{bogoslavskyi2016FastRangeImagebased}
            & range image & given RI & no & $\Delta t_F{+}\Delta t_P$ \\
        TRAVEL~\cite{oh2022TRAVELTraversableGround}
            & scan graph & org.\ scan & no & $\Delta t_F{+}\Delta t_P$ \\
        Pi-Lisco~\cite{najdataei2022PiLiscoParallelIncremental}
            & range image & org.\ scan & yes & $\Delta t_C{+}\Delta t_P$ \\
        Reich~\cite{reich2024LowLatencyInstance}
            & range image & org.\ scan & yes & $\Delta t_C{+}\Delta t_P$ \\
        \mbox{C-ARC}
            & range grid & adaptive & yes & $T_{\mathrm{pub}}{+}\Delta t_P$\\
        \bottomrule
        \multicolumn{5}{@{}p{\dimexpr\columnwidth-2\tabcolsep\relax}@{}}{%
        \footnotesize\textsuperscript{*}Layout: source of $H{\times}W$---
        given RI / org.\ scan from a formed range image or manufacturer scan structure (PBID, beams, sweep);
        adaptive: FOV-aspect $H{\times}W$ from a short $T_B$ window with $M_\mu\le T_M$ (no scan layout).
        Cont.: continuous sliding-window clustering (vs.\ discrete frames).
        Publish (after an instance is observable):
        $\Delta t_F$ frame period;
        $\Delta t_C$ time until the spinning sweep has covered the instance;
        $T_{\mathrm{pub}}{=}1/f_{\mathrm{pub}}$ application poll;
        $\Delta t_P$ processing.}
    \end{tabularx}
\end{table}

\noindent \mbox{C-ARC} addresses these limitations with an adaptive FOV-aspect range grid, a two-layer graph over a sliding window, and poll-timed retrieval at $T_{\mathrm{pub}}$.


\section{METHOD}
\label{sec:evaluation}
\noindent \mbox{C-ARC} is a continuous clustering framework that maintains a persistent connectivity graph over a sliding point window $T_B$, set by the user rather than by a hardware frame period $\Delta t_\mathrm{F}$.
A typical $\Delta t_\mathrm{F}$ of ${\sim}$50--100\,ms is already temporally mixed under ego-motion and moving objects; $T_B$ simply makes that window length an explicit design choice.

Incoming points are projected onto a bucket-based range grid, enabling neighbourhood search in $O(w^2k)$, where $k$ denotes the number of points per cell and $w$ the diameter of the search window, while remaining independent of the sensor scanning pattern.
This design addresses two key challenges of non-repetitive LiDAR sampling:

\begin{enumerate}
	\item \textbf{Irregular Spatial Coverage:} Non-repetitive trajectories do not fill the FOV uniformly, so some angular cells receive several returns while others stay empty within a given $T_B$.
	      Our bucket range grid retains the surplus returns that an overwrite range image would lose (Sec.~II), without a manufacturer scan layout.
	\item \textbf{Ill-defined Scan Boundaries:} Without a periodic scan cycle, frame boundaries become heuristic.
	      Longer accumulation windows improve coverage but increase latency, while shorter windows reduce latency at the cost of structural completeness.
\end{enumerate}

Choosing $H{\times}W$ relative to $T_B$ still forms a \textit{Sparsity-Collision} trade-off: too fine a grid relative to the buffer leaves many cells empty and weak neighbourhood support; too coarse a grid increases collisions per cell.

\subsection{Adaptive Range Grid Sizing}\label{sec:adaptive_range_grid}
\noindent Because the optimal $T_B$ depends on application-specific latency and density requirements, \mbox{C-ARC} treats it as a configurable parameter rather than a fixed design choice.
Latency-critical tasks favour short $T_B$ and coarser grids; mapping can use longer windows and finer bins.
Non-repetitive sensors provide no manufacturer $H{\times}W$, so the grid must be inferred from the sensor FOV and a short start-of-stream window of length~$T_B$.

We choose an angular cell size $r$ with FOV aspect so that a square \(w{\times}w\) neighbourhood in $(u,v)$ spans a comparable solid angle in elevation and azimuth:
$H = \lfloor\mathrm{FOV}_v/r \rceil$,
$W = \lfloor\mathrm{FOV}_h/r\rceil$,
which enforces $W/H \approx \mathrm{FOV}_h/\mathrm{FOV}_v$.

Let $\mathbf{F} \in \mathbb{N}_0^{H \times W}$ count points per cell and $\mathcal{V} = \{ (u,v) \mid F_{u,v} > 0 \}$.
Mean multiplicity
$M_\mu = \frac{1}{|\mathcal{V}|}\sum_{(u,v)\in\mathcal{V}} F_{u,v}$
measures collisions under the chosen~$r$.
A $w{\times}w$ neighbourhood aggregates the points in its cells; occupied cells hold on average $M_\mu$ points, so $M_\mu$ directly controls the cost of each local search.

\textsc{AdaptiveGridInit} selects the coarsest admissible cell size, $r^\star = \max\{\, r \mid M_\mu(r) \le T_M \,\}$; finer grids also respect the threshold but spend cells on empty bins and weaken neighbourhood support.
Since $M_\mu$ increases with~$r$, we search by bisection below a coarse bound of $1^\circ$, re-projecting the initialization window at each trial, and stop when $M_\mu$ is within $\delta{=}0.15$ of $T_M$ from below or further search no longer changes~$r$. 
$T_M$ is a multiplicity threshold chosen \emph{a priori} rather than fit to segmentation scores; we use $T_M = 2$ unless stated.
Using mean occupancy $M_\mu$ enforces a global resolution matched to average FOV density. 
Sizing the grid to local occupancies (e.g. Livox Avia's concentrated centre) would over-refine sparse peripheral regions.
The resulting $(H,W)$ follows from the sensor FOV, its sampling pattern and~$T_B$, and is estimated once on an initialization window and held for the sequence.

\subsection{C-ARC Architecture}
\label{sec:carc_architecture}
\noindent \mbox{C-ARC} treats point cloud clustering as a dynamic connectivity graph problem, where points are vertices and spatial adjacency defines edges.
Unlike frame-based methods that reset state after each scan, \mbox{C-ARC} maintains a persistent graph over time.
Modern LiDAR sensors such as the Livox Mid-360 stream up to 200~kHz, while downstream applications typically operate at 10--20~Hz, making full-rate clustering computationally infeasible.
To address this, \mbox{C-ARC} decouples point insertion from cluster retrieval: insertion runs continuously at sensor rate, while retrieval is executed on demand at application-specific frequencies.
Algorithm~\ref{alg:carc} summarizes the resulting loop.

\begin{algorithm}[t]
	\caption{C-ARC Main Loop}\label{alg:carc}
	\begin{algorithmic}[1]
		\Require Point stream $\mathcal{P}$, sliding window
		$T_B$, load threshold $T_M$, mask size $w$,
		sensor FOV
		\Ensure Labelled point set $\{(\boldsymbol{p}_i, \ell_i)\}$
		\State Initialize ring buffer $B$, range grid $S$,
		graphs $G_P$, $G_C$
		\State $t_0 \leftarrow \textsc{Now}()$
		\While{$\textsc{Now}() - t_0 < T_B$}
		\State $B.\textsc{Insert}(\mathcal{P}.\textsc{Next}())$
		\Comment{prime buffer to capacity}
		\EndWhile
		\State $S \leftarrow \textsc{AdaptiveGridInit}(B,\, T_M)$
		\Comment{Sec.~\ref{sec:adaptive_range_grid}}
		\State $\textsc{InsertAll}(B,\, S,\, G_P,\, G_C,\, w)$
        \Comment{sizing window reused; first output after $T_B$}

		\While{$\mathcal{P}$ is streaming}
		\State $\boldsymbol{p}_\mathrm{new} \leftarrow
			\mathcal{P}.\textsc{Next}()$
		\If{$B.\textsc{IsFull}()$}
		\State \textsc{Evict}$(B.\textsc{Oldest}(),\,
			S,\, G_P,\, G_C)$
		\Comment{Remove point; mark affected clusters
			\textit{pending}}
		\EndIf
		\State \textsc{Insert}$(\boldsymbol{p}_\mathrm{new},\,
			B,\, S,\, G_P,\, G_C,\, w)$
		\Comment{Fig.~\ref{fig:point_insertion}}
		\If{\textsc{PublicationDue}()}
		\State \textsc{ValidatePending}$(G_P)$
		\Comment{Fig.~\ref{fig:cc_retrieval:pre_validation}}
		\State \textsc{UpdateComponentGraph}$(G_C)$
		\Comment{Fig.~\ref{fig:cc_retrieval:post_validation}}
		\State \textbf{yield} $\textsc{BFS}(G_C)
			\rightarrow \{(\boldsymbol{p}_i, \ell_i)\}$
		\Comment{Fig.~\ref{fig:cc_retrieval:graph_c_bfs}}
		\EndIf
		\EndWhile
	\end{algorithmic}
\end{algorithm}

\begin{figure*}[t]
	\centering
	\includegraphics[width=0.9\textwidth]{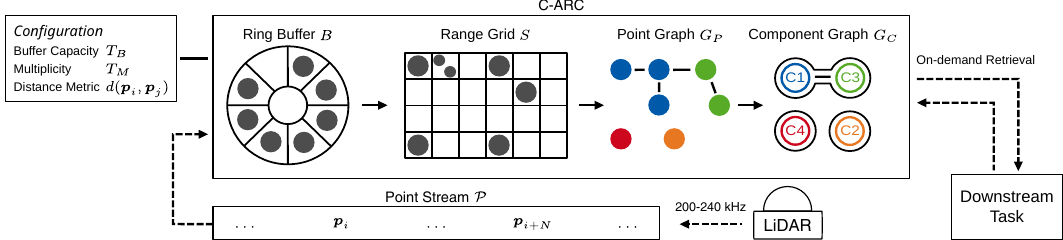}
	\caption{The \mbox{C-ARC} pipeline. Incoming points are stored Rin a ring buffer and indexed via a FOV-sized range grid. The dual-graph structure decouples high-frequency insertion (200--240~kHz) from lower-frequency cluster retrieval. The downstream-task block illustrates potential use of the resulting clusters and is not evaluated in this work.}\label{fig:c_arc_architecture}
\end{figure*}

Fig.~\ref{fig:c_arc_architecture} illustrates how the pipeline components interact.
Accumulating points this way uncouples output latency from $T_B$.
Processing relies on three core data structures:

\noindent
\textbf{(i)~Ring Buffer:} a count-based FIFO window stores the $N = T_B \cdot r_s$ most recent points for  sampling rate $r_s$ (eviction is strictly count-driven to ensure $O(1)$ push/pop overhead); 
\textbf{(ii)~Range Grid:} the spatial index supporting neighbourhood queries; and 
\textbf{(iii)~Dual-Graph Hierarchy:} a point-level adjacency graph $G_P$ forming local clusters, and a component graph $G_C$ whose nodes are local clusters and edges are inter-cluster bridges forming connected components (CCs).

\noindent Two-layer graphs appear in frame-based range/set clustering~\cite{yang2020TwoLayerGraphClusteringRealTime} and continuous point-tree clustering~\cite{reich2024LowLatencyInstance}.
\mbox{C-ARC} builds upon the idea of linking connected components rather than merging their point sets~\cite{reich2024LowLatencyInstance} by representing the relationship between local structures through an edge in the high-level graph ($G_C$). 
However, unlike \cite{reich2024LowLatencyInstance}, \mbox{C-ARC} also maintains undirected edges between points belonging to different local structures in the point-level graph ($G_P$). 
This additional connectivity is required because \mbox{C-ARC} operates on a sliding window in which points may be removed over time, potentially invalidating existing connections and connected components.
To maintain both graphs under point removal, the system performs three operations.

\subsubsection{Incremental Point Insertion}
\label{sec:inc_point_insertion}
\noindent Upon arrival, point $\boldsymbol{p}_i$ is appended to the ring buffer $B$ and projected onto grid coordinates $(u_i, v_i)$ using a standard spherical projection.
We then perform a neighbourhood search using a $w \times w$ spatial mask $\mathcal{M}_w$ centred at the projection coordinates.
A candidate point $\boldsymbol{p}_j$ is accepted as a neighbour if it satisfies the configured distance metric
$\mathcal{N}(\boldsymbol{p}_i) = \{ \boldsymbol{p}_j \in \mathcal{M}_w(\boldsymbol{p}_i) \setminus \{\boldsymbol{p}_i\} \mid d(\boldsymbol{p}_i, \boldsymbol{p}_j) \le d_\mathrm{max} \}$.

\begin{figure}[!t]
	\centerline{
		\subfigure[Isolation]{%
			\includegraphics[width=0.33\columnwidth]{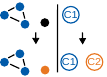}\label{fig:point_insertion:isolation}
		}%
		\hfil
		\subfigure[Expansion]{%
			\includegraphics[width=0.33\columnwidth]{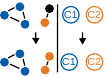}\label{fig:point_insertion:expansion}
		}%
		\hfil
		\subfigure[Bridging]{%
			\includegraphics[width=0.33\columnwidth]{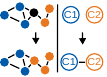}\label{fig:point_insertion:bridging}
		}%
	}
	\caption{Insertion of a new point (black). Updates to $G_P$ (left) and $G_C$ (right).
		(a)~Isolation: no neighbours found; a new local cluster is created.
		(b)~Expansion: all neighbours belong to one cluster; the point joins it and edges are added to $G_P$.
		(c)~Bridging: neighbours span multiple clusters; an edge is added to $G_C$ to connect them without merging the local point sets.}\label{fig:point_insertion}
	\medskip
	\setcounter{subfigure}{0}
	\centerline{
		\subfigure[Validating clusters]{%
			\includegraphics[width=0.33\columnwidth]{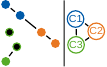}\label{fig:cc_retrieval:pre_validation}
		}%
		\hfil
		\subfigure[Update $G_C$]{%
			\includegraphics[width=0.33\columnwidth]{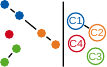}\label{fig:cc_retrieval:post_validation}
		}%
		\hfil
		\subfigure[BFS on CCs]{%
			\includegraphics[width=0.33\columnwidth]{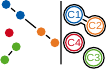}\label{fig:cc_retrieval:graph_c_bfs}
		}%
	}
	\caption{On-demand retrieval process.
		(a)~Pending clusters (green) are revalidated using BFS to restore structural consistency after point removals, updating the out-of-sync graph $G_C$.
		(b)~The component graph $G_C$ is synchronized with the validated cluster structure.
		(c)~A final BFS over $G_C$ extracts all CCs and assigns each buffered point the label of its corresponding component.}\label{fig:cc_retrieval}
\end{figure}

\noindent Since \mbox{C-ARC} is metric-agnostic, $d(\cdot, \cdot)$ may be any single-linkage distance criterion defined on point pairs.
Depending on the neighbourhood, insertion is an \textit{isolation} case ($\mathcal{N}(\boldsymbol{p}_i) = \emptyset$), an \textit{expansion} case, or a \textit{bridging} case, as detailed in Fig.~\ref{fig:point_insertion}.
When bridging, $\boldsymbol{p}_i$ is assigned to the most recently created of the spanned clusters to maximize the temporal longevity of the connection, and those clusters are joined by a high-level edge in $G_C$ instead of merging their point sets.

\subsubsection{Point Removal and Lazy Deletion}
\noindent When a point is evicted from the ring buffer, it is removed from the grid, and its edges are deleted from $G_P$.
Validating whether the removal splits a cluster is computationally too expensive at sensor sampling rates.
To minimize redundant traversals, \mbox{C-ARC} employs lazy deletion.
Rather than immediately validating each removal, affected clusters are marked as \enquote{pending}.
When consecutive points from the same cluster are evicted, the cluster remains in the pending state throughout, deferring all BFS validation until the next retrieval.
This amortizes what would otherwise be many individual traversals into a single pass.

\subsubsection{Connected Component Retrieval}
\noindent Triggered on demand (typically at fixed $f_{\mathrm{pub}}$), retrieval extracts a snapshot of the dynamic graph in three steps (Fig.~\ref{fig:cc_retrieval}): every \enquote{pending} cluster is validated by BFS to detect splits caused by point removals, the bridge edges of $G_C$ are synchronized with the validated local clusters, and a final BFS over $G_C$ identifies the CCs, labelling each of the $N$ buffered points with its component.

\noindent A simpler alternative at the same $T_B$ is to buffer points and rebuild the full sliding window on each poll (batch-rebuild baseline; Sec.~\ref{sec:computational_analysis}).
\mbox{C-ARC} instead applies the insertion update of Alg.~\ref{alg:carc} after each incoming packet, so retrieval validates and exports labels from a state that already reflects the current buffer, rather than reconstructing connectivity from scratch on every poll.

\section{EVALUATION}

\subsection{Experimental Setup}\label{sec:experimental_setup}
\noindent \mbox{C-ARC} is implemented as a stand-alone \mbox{C++17} library.
Experiments use offline replay of recorded sequences on a machine with an
Intel\textsuperscript{\textregistered} Core\textsuperscript{\texttrademark} i7-1280P CPU (up to 4.8 GHz) and 32 GB RAM, running Ubuntu 24.04 LTS.
Real-time feasibility is assessed by comparing measured processing and poll-cycle times with the specified publication-rate budgets.
The application was executed single-threaded.
Because no public dataset combines diverse non-repetitive sensors, spinning-LiDAR baselines, and cluster labels, each benchmark in Tab.~\ref{tab:datasets} is used for a distinct evaluation axis (scenes and sequences in the table footnote).
Under- and over-segmentation entropy (USE/OSE)~\cite{oh2022TRAVELTraversableGround} follow the clustering protocol of Reich~\cite{reich2024LowLatencyInstance}, with semantic-derived instance labels as a proxy.

\begin{table}[!t]
    \caption{Overview of evaluation datasets.}
    \label{tab:datasets}
    \centering
    \footnotesize
    \setlength{\tabcolsep}{4pt}
    \begin{tabularx}{\columnwidth}{@{}>{\raggedright\arraybackslash\hsize=1.05\hsize}X >{\raggedright\arraybackslash\hsize=0.95\hsize}X >{\raggedright\arraybackslash\hsize=1.0\hsize}X@{}}
        \toprule
        \textbf{Dataset} & \textbf{Sensor} & \textbf{Role} \\
        \midrule
        \textit{S.MID}~\cite{wangSFPNetSparseFocal2025}
                         & Livox Mid-360
                         & OSE/USE non-rep. \\
        \textit{semanticKITTI}~\cite{behleySemanticKITTIDatasetSemantic2019}
                         & Velodyne HDL-64E
                         & OSE/USE rep. \\   
         \textit{HK-MEMS}~\cite{ruanHKMEMSMEMSLiDAR2024}
                         & RoboSense M1
                         & adaptive sizing only \\
        \textit{Multi-modal}~\cite{multi_modal_lidar_dataset}
                         & Livox \mbox{Mid-360}, Avia
                         & Latency + qualitative \\
                         & Ouster OS0-128
                         & adaptive sizing only \\
        \textit{Rep-or-not}~\cite{xie2024NonRepetitivePromisingLiDAR}
                         & Livox HAP (CARLA)
                         & qualitative only \\
        \bottomrule
    \end{tabularx}\par
    \vspace{4pt}
    {\footnotesize\raggedright
    Multi-modal: IndoorOffice1, OutdoorRoad, OutdoorForest.
    S.MID: v1.3, sequence~02.
    SemanticKITTI: sequences~00--10.
    Rep-or-not: urban, one scan.
    rep.: repetitive\par}
\end{table}

\subsection{Segmentation quality}
\label{sec:segmentation_quality}


\noindent We evaluate segmentation quality on S.MID and SemanticKITTI, covering both non-repetitive and organised LiDAR data.

\noindent\textbf{S.MID.}
On S.MID sequence~02 we compare overwrite-based range image clustering~\cite{bogoslavskyi2016FastRangeImagebased} to \mbox{C-ARC}'s bucket range grid under a matched protocol: beam-angle criterion $\theta{=}10^\circ$, neighbourhood $w{=}5$ (Chebyshev radius $r{=}2$), and spherical layout $H{\times}W{=}73{\times}444$ from adaptive sizing on that sequence at $T_B{=}0.1$\,s.
Both methods are evaluated on every labelled Mid-360 frame of the sequence ($n{=}5000$), with instance labels derived from the dataset semantics as proxy ground truth.
Tab.~\ref{tab:smid_use_ose} therefore isolates the projection model (bucket vs.\ overwrite) rather than differences in threshold, neighbourhood, or grid size.
At this fine layout, collisions remain mild and the USE/OSE gap is accordingly small.
\newcommand{\gridcompimg}[1]{%
    \includegraphics[trim=0pt 0pt 0pt 0pt,clip,width=0.24\linewidth]{#1}}
\begin{figure*}[t]
    \centerline{
        \subfigure[Baseline: ($177 \times 803$)]{%
            \gridcompimg{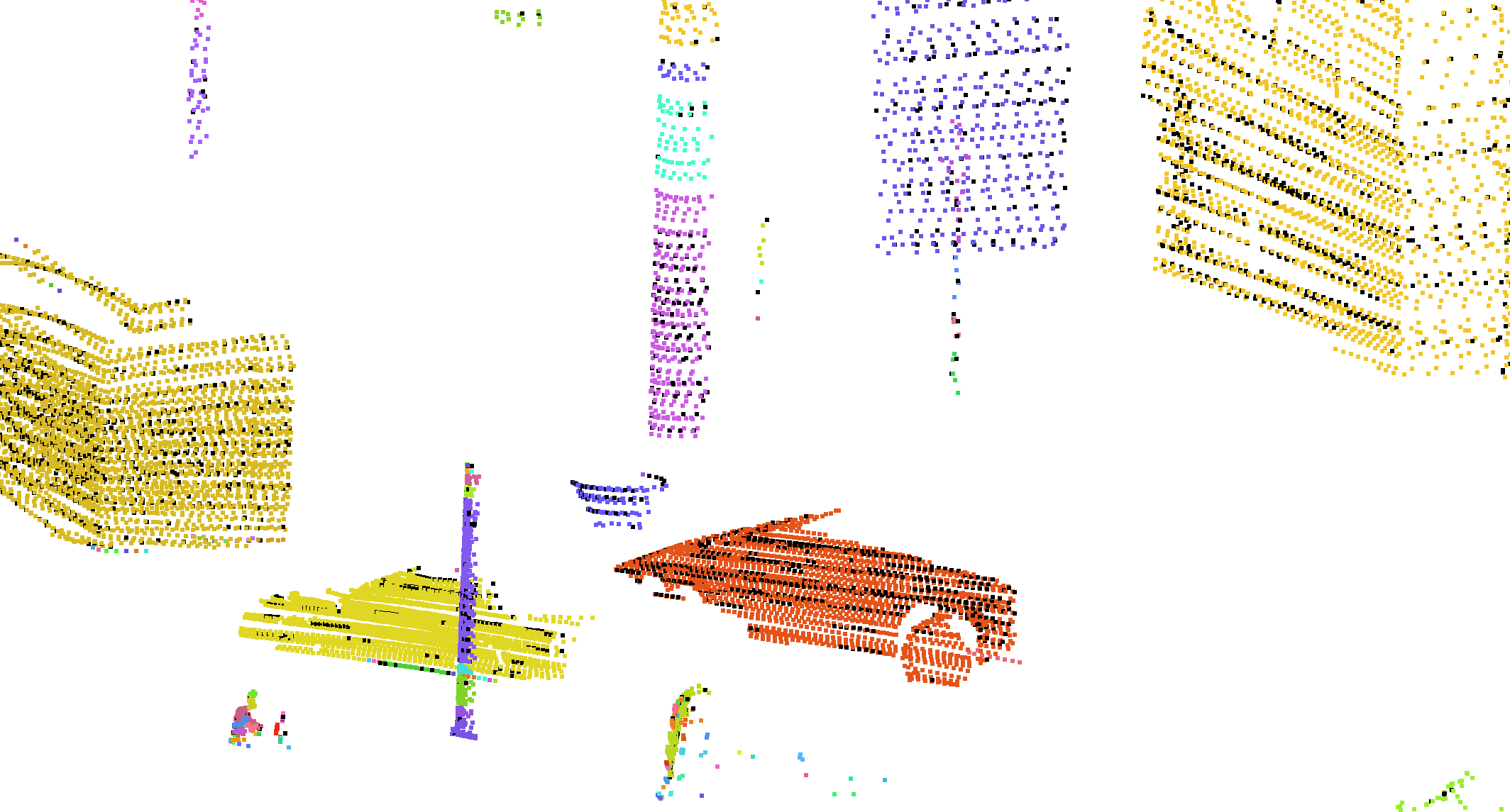}%
            \label{fig:grid_resolution_comp:a}%
        }\hfil
        \subfigure[\mbox{C-ARC}: ($177 \times 803$)]{%
            \gridcompimg{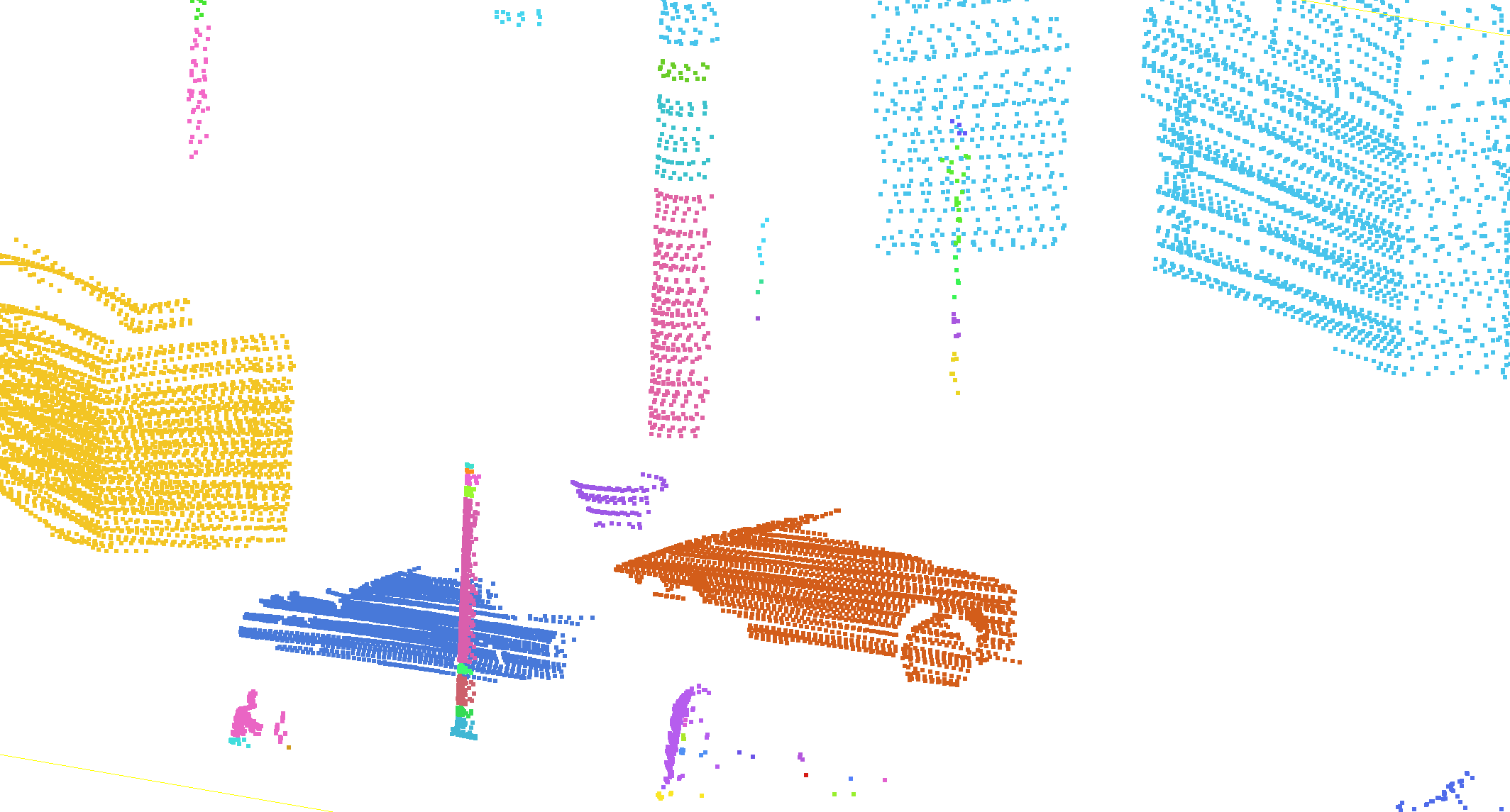}%
            \label{fig:grid_resolution_comp:b}%
        }\hfil
        \subfigure[Baseline: ($94 \times 427$)]{%
            \gridcompimg{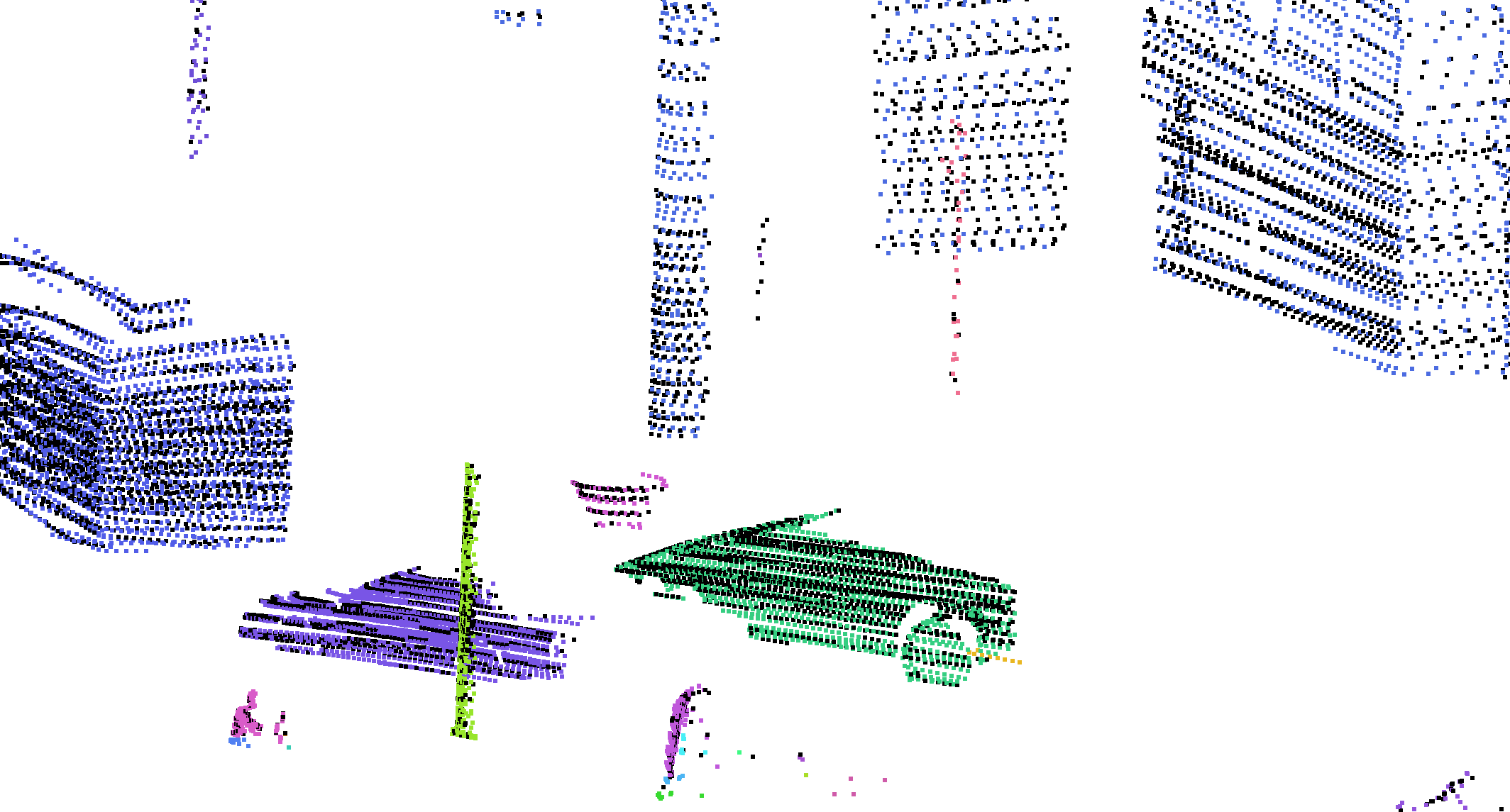}%
            \label{fig:grid_resolution_comp:c}%
        }\hfil
        \subfigure[\mbox{C-ARC}: ($94 \times 427$)]{%
            \gridcompimg{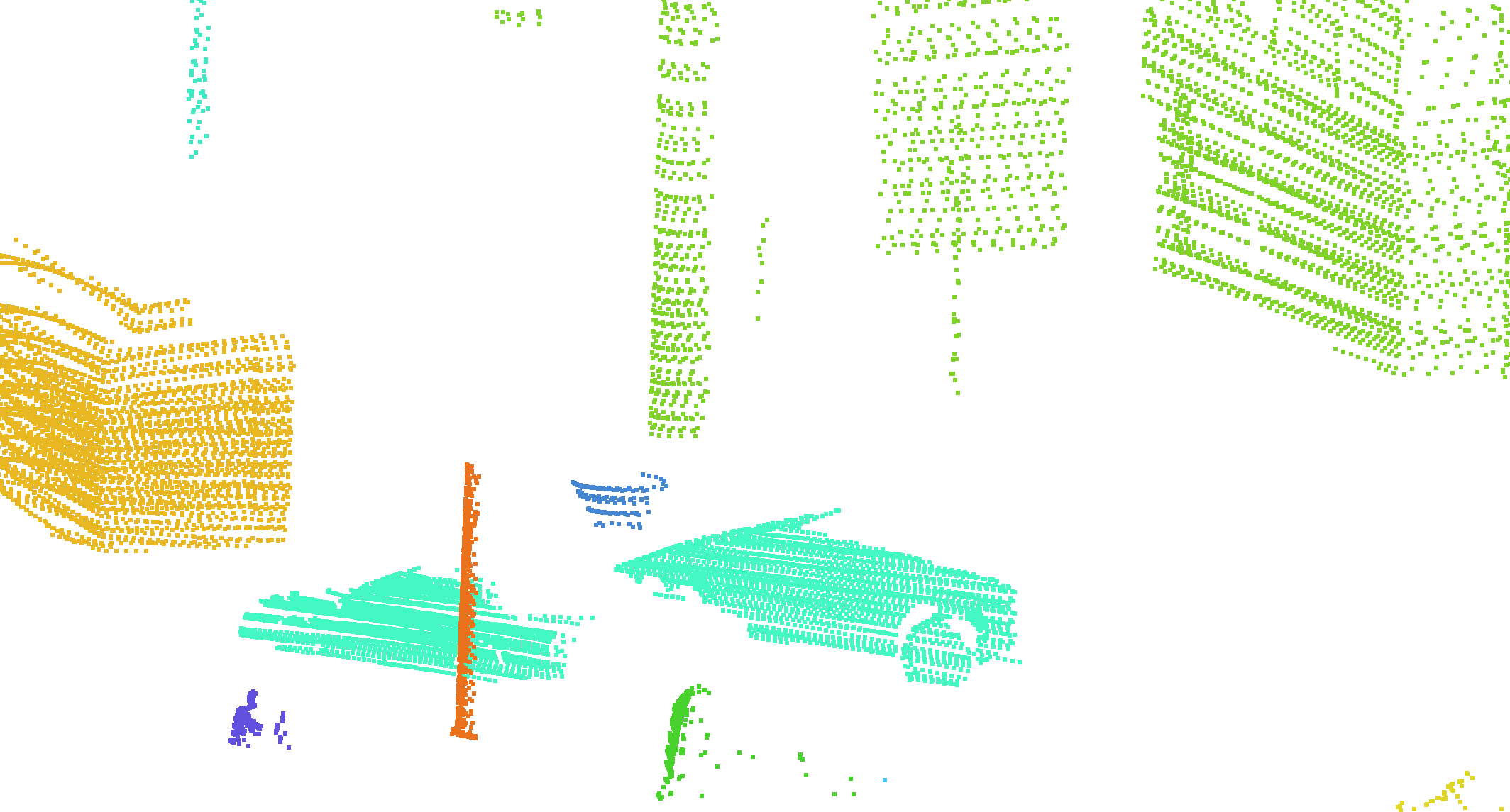}%
            \label{fig:grid_resolution_comp:d}%
        }
    }
    \caption{Qualitative comparison of overwrite-based range image clustering~\cite{bogoslavskyi2016FastRangeImagebased} and \mbox{C-ARC}'s bucket-based range grid on a non-repetitive LiDAR scan~\cite{xie2024NonRepetitivePromisingLiDAR}.
Fine-resolution projections ($M_\mu = 1.25$; \subref{fig:grid_resolution_comp:a},\subref{fig:grid_resolution_comp:b}) reduce collisions (black pixels) but produce fragmentation in the baseline due to sparse projections.
Coarse projections ($M_\mu = 2$) improve connectivity but introduce information loss \subref{fig:grid_resolution_comp:c}.
By accumulating multiple points per cell, \mbox{C-ARC} preserves structural connectivity without information loss \subref{fig:grid_resolution_comp:d}.
All experiments use $w = 5$ and $\theta = 10^\circ$.
Ground points are filtered for visual clarity.}
\label{fig:grid_resolution_comp}
\end{figure*}

\begin{table}[!t]
    \caption{USE/OSE on S.MID sequence~02.
    Matched protocol: $\theta{=}10^\circ$, $w{=}5$, $H{\times}W{=}73{\times}444$.}
    \label{tab:smid_use_ose}
    \centering
    \footnotesize
    \setlength{\tabcolsep}{4pt}
    \begin{tabular}{@{}lcc@{}}
        \toprule
        \textbf{Method} & \textbf{USE}~$\mu{\downarrow}/\sigma{\downarrow}$ & \textbf{OSE}~$\mu{\downarrow}/\sigma{\downarrow}$ \\
        \midrule
        Bogoslavskyi~\cite{bogoslavskyi2016FastRangeImagebased} & 16.92 / 7.57 & 20.57 / 10.39 \\
        \mbox{C-ARC} & 12.63 / 6.05 & 18.09 / 9.38 \\
        \bottomrule
    \end{tabular}
\end{table}

\noindent\textbf{SemanticKITTI.}
TRAVEL~\cite{oh2022TRAVELTraversableGround} and Reich~\cite{reich2024LowLatencyInstance} target organized spinning-LiDAR scans; we report their cited USE/OSE~\cite{oh2022TRAVELTraversableGround} on sequences~0--10 with Euclidean instance labels derived from semantic annotations ($d_T = 0.5$\,m) and evaluate \mbox{C-ARC} under the same labels and metric.

\noindent To report USE/OSE on that split, \mbox{C-ARC} is run outside its default operating mode.
Points are first filtered with the column-wise slope ground removal of Reich~\cite{reich2024LowLatencyInstance}, applied in preprocessing only and not part of the published pipeline.
Each scan is projected onto Reich's fixed $64 \times 2200$ range grid and clustered with the settings of their released SemanticKITTI evaluation (Euclidean distance $d_T = 0.5$\,m, at most 20 search steps per axis), here a Chebyshev neighbourhood radius $r{=}20$,
and ingested and published once after a full $360^{\circ}$ sweep.
This yields a single label set per scan rather than the continuous, on-demand retrieval at $f_{\mathrm{pub}}$ from Sec.~III.
Hence, the aim is not optimality on mechanical LiDAR, but to show that segmentation quality remains in the same range as methods built for that sensor class.
Continuous operation under the default retrieval policy is evaluated separately in Sec.~\ref{sec:computational_analysis}.
Tab.~\ref{tab:skitti_use_ose} summarizes the results.

\begin{table}[!t]
    \caption{USE/OSE on SemanticKITTI (seq.~0--10); \mbox{C-ARC} run in non-continuous mode (see text).}
    \label{tab:skitti_use_ose}
    \centering
    \footnotesize
    \setlength{\tabcolsep}{4pt}
    \begin{tabular}{@{}lcc@{}}
        \toprule
        \textbf{Method} & \textbf{USE}~$\mu{\downarrow}/\sigma{\downarrow}$ & \textbf{OSE}~$\mu{\downarrow}/\sigma{\downarrow}$ \\
        \midrule
        TRAVEL~\cite{oh2022TRAVELTraversableGround} & 24.07 / 11.88 & 70.40 / 34.44 \\
        Reich~\cite{reich2024LowLatencyInstance} & 23.74 / 11.21 & 65.23 / 23.88 \\ 
        \mbox{C-ARC} & 21.60 / 6.26 & 47.78 / 8.97 \\
        \bottomrule
    \end{tabular}
\end{table}

\subsection{Adaptive range grid sizing}
\label{sec:adaptive_range_grid_eval}

\noindent Non-repetitive streams exacerbate point collisions in overwrite-based range images (Sec.~II).
We apply \textsc{AdaptiveGridInit} independently to each sensor in Tab.~\ref{tab:grid_init}, covering the Mid-360 and Avia, RoboSense M1, and repetitive LiDARs from Ouster and Velodyne.
The resulting $H{\times}W$ adapts to the sensor FOV and sampling characteristics while maintaining $M_\mu$ close to the target occupancy.
For the same sensor, increasing $T_B$ produces a finer grid at approximately the same FOV aspect, as more points cause a smaller angular cell size before the occupancy constraint is reached.
The objective is not to reproduce manufacturer line counts, but to show that one control rule adapts the grid to substantially different sensing configurations.

\begin{table}[!t]
    \caption{Adaptive range grid sizing ($T_M{=}2$).
    $M_\mu$ is the occupied-cell mean from Sec.~III-A.}
    \label{tab:grid_init}
    \centering
    \footnotesize
    \setlength{\tabcolsep}{3pt}
    \begin{tabular}{@{}l c c c@{}}
      \toprule
      \textbf{Sensor} & $T_B$ & $H{\times}W$ & $M_\mu$ \\
      \midrule
      Mid-360 & 0.1 & $47{\times}287$ & 1.90 \\
      Mid-360 & 0.3 & $87{\times}531$ & 1.89 \\
      Avia & 0.1 & $169{\times}154$ & 1.90 \\
      Avia & 0.3 & $272{\times}248$ & 1.86 \\
      RoboSense M1 & 0.1 & $185{\times}888$ & 1.97 \\
      Ouster OS0-128 & 0.1 & $168{\times}671$ & 1.86 \\
      Velodyne HDL-64E & 0.1 & $161{\times}2159$ & 1.97 \\
      \bottomrule
    \end{tabular}
\end{table}

\noindent\textbf{Range image vs.\ range grid.}
Fig.~\ref{fig:grid_resolution_comp} isolates the projection model on a single non-repetitive scan~\cite{xie2024NonRepetitivePromisingLiDAR}.
Here, $M_\mu$ is only a \emph{diagnostic} of mean cell occupancy, not a segmentation score.
For overwrite projections, $M_\mu > 1$ means information loss ($M_\mu \approx 2$ implies roughly every second point overwrites a prior observation), whereas \mbox{C-ARC} decouples resolution from collision rate and can use a coarser $H \times W$ without discarding returns.

\noindent\textbf{Grid policy on S.MID.}
Tab.~\ref{tab:grid_ablation} compares the proposed adaptive grid with fixed-resolution alternatives on S.MID (Livox Mid-360), using otherwise identical \mbox{C-ARC} settings.
The fixed fine grid uses the same FOV aspect as the adaptive grid, with half the cell area.
The fixed-best grid is selected from the same aspect-preserving scale range using USE$+$OSE on a held-out stride subset.

\begin{table}[!t]
    \caption{Grid-policy ablation on S.MID Mid-360 \\($T_B{=}0.1\,\mathrm{s}$, $\theta{=}10^\circ$, $w{=}5$).}
    \label{tab:grid_ablation}
    \centering
    \footnotesize
    \setlength{\tabcolsep}{3pt}
    \begin{tabularx}{\columnwidth}{@{}>{\raggedright\arraybackslash}X c c c c@{}}
        \toprule
        \textbf{Policy} & $H{\times}W$ & $M_\mu$ & \textbf{USE}~$\mu{\downarrow}/\sigma{\downarrow}$ & \textbf{OSE}~$\mu{\downarrow}/\sigma{\downarrow}$ \\
        \midrule
        Adaptive (proposed) & $73{\times}444$ & 1.90 & 12.63 / 6.05 & 18.09 / 9.38 \\
        Fixed fine & $103{\times}628$ & 1.43 & 11.19 / 5.75 & 36.41 / 18.01 \\
        Fixed best (search) & $36{\times}220$ & 5.06 & 9.79 / 5.33 & 2.98 / 2.44 \\
        Overwrite~\cite{bogoslavskyi2016FastRangeImagebased} & $73{\times}444$ & 1.90 & 16.92 / 7.57 & 20.57 / 10.39 \\
        \bottomrule
    \end{tabularx}
\end{table}

\noindent Across the three resolutions, coarsening improves both scores: larger cells offer more neighbour candidates per query and fragment objects less, so USE and OSE fall together, and the searched fixed-best grid attains the lowest values of both.
That optimum is not free: at $M_\mu{\approx}5$ it holds ${\sim}2.6\times$ the points per occupied cell, and neighbourhood search scales with occupancy (Sec.~III), which is what $T_M$ bounds.
Adaptive sizing therefore trades a tuned quality optimum for a prescribed cost level, without per-sensor grid search.

\subsection{Computational Analysis}
\label{sec:computational_analysis}
\begin{figure*}[!t]
    \centering
    \subfigure[$T_B = 0.1\,\mathrm{s}$, ($47\times 287$)]{%
        \includegraphics[width=0.24\linewidth]{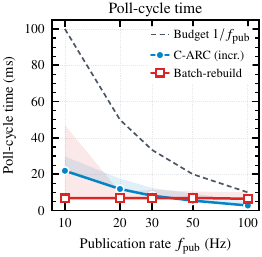}
        \label{fig:inc_vs_batch_a}
    }\hfil
    \subfigure[$T_B = 0.3\,\mathrm{s}$, ($87\times 531$)]{%
    \includegraphics[width=0.24\linewidth]{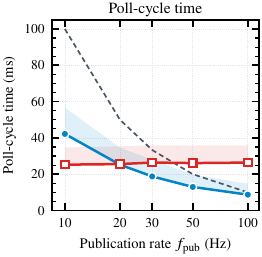}
        \label{fig:inc_vs_batch_b}
    }\hfil
    \subfigure[$T_B = 0.3\,\mathrm{s}$, $f_\mathrm{pub} = 50\,\mathrm{Hz}$]{%
    \includegraphics[width=0.24\linewidth]{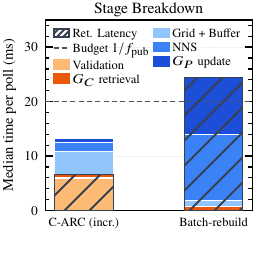}
        \label{fig:inc_vs_batch_stages}
    }\hfil
    \subfigure[$T_B = 0.3\,\mathrm{s}$]{%
    \includegraphics[width=0.24\linewidth]{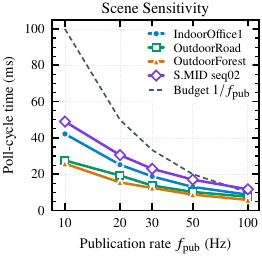}
        \label{fig:inc_vs_batch_scene}
    }\hfil
    \caption{
    \mbox{C-ARC} vs.\ batch-rebuild (Livox Mid‑360; IndoorOffice1~\cite{multi_modal_lidar_dataset} unless stated). (\subref{fig:inc_vs_batch_a},\subref{fig:inc_vs_batch_b}) Poll‑cycle time vs. publication rate $f_{\mathrm{pub}}$ at $T_B=0.1$ s and $0.3$ s (median; shaded: median--$P_{99}$). (c) Stage breakdown at $f_{\mathrm{pub}}=50$ Hz for (b): stacked bars = per‑stage median contributions (see Sec.~\ref{sec:carc_architecture}); hatched overlay = retrieval latency. (d) Scene sensitivity: C‑ARC median poll‑cycle time at $T_B=0.3$ s across four Mid‑360 sequences.}
    \label{fig:inc_vs_batch}
\end{figure*}
\noindent\textbf{Incremental vs batch-rebuild.}
We compare incremental \mbox{C-ARC} against a \emph{batch-rebuild baseline} that reconstructs projection, clustering, and labelling from scratch at each poll (Sec.~III).
Incremental keeps $G_P$ current after every packet, whereas batch-rebuild concentrates the same work at the poll.
The metric is \emph{poll-cycle time}: for \mbox{C-ARC}, accumulated insertion since the last poll plus validation and retrieval; for batch-rebuild, full-window rebuild plus retrieval.
Both must stay below $1/f_{\mathrm{pub}}$ for real-time output at rate~$f_{\mathrm{pub}}$.
Fig.~\ref{fig:inc_vs_batch} reports median and $P_{99}$ on Livox Mid-360 IndoorOffice1 (further scenes in Fig.~\ref{fig:inc_vs_batch_scene}); budget claims use the median.
Both methods share one grid per~$T_B$, estimated on the IndoorOffice1 window at $T_M{=}2$ ($47{\times}287$ at $T_B{=}0.1$\,s, $87{\times}531$ at $T_B{=}0.3$\,s) and reused for every sequence, so the comparison varies only in scene.

Incremental insertion is paid once per point and is independent of $f_{\mathrm{pub}}$, so \mbox{C-ARC} poll cycles shorten with $f_{\mathrm{pub}}$ while batch-rebuild pays the full window at every poll and stays flat (Fig.~\ref{fig:inc_vs_batch_a}).
The curves cross, and the crossing moves to lower rates as $T_B$ grows (${\sim}40$~Hz at $T_B{=}0.1$~s, ${\sim}20$~Hz at $T_B{=}0.3$~s).
With increasing frequency, only \mbox{C-ARC} fits the budget, because the poll only validates and reads $G_C$ instead of re-clustering the entire $T_B$ buffer (Fig.~\ref{fig:inc_vs_batch_b},~\ref{fig:inc_vs_batch_stages}).

Poll-cycle time includes insertion work performed between polls. Upon a poll request, retrieval latency is the hatched portion in Fig.~\ref{fig:inc_vs_batch_stages}: ${\sim}12.5$~ms for \mbox{C-ARC}, comprising only $G_C$ retrieval and validation, versus ${\sim}24$~ms for batch-rebuild, which must re-project and re-cluster the current $T_B$ buffer from scratch.

\noindent\textbf{Back-to-back throughput.}
As a complementary throughput bound, we measure \emph{back-to-back polling}: the next retrieval starts as soon as the previous one finishes ($f_{\max}{\approx}1/T_{\mathrm{cycle}}$).
Tab.~\ref{tab:fmax} reports median $f_{\max}$ on four Mid-360 sequences.
At $T_B{=}0.1$~s, \mbox{C-ARC} achieves the higher poll rate on all sequences.
At $T_B{=}0.3$~s, the gap narrows and depends on scene composition: \mbox{C-ARC} remains faster on IndoorOffice1 and OutdoorForest, while batch-rebuild is faster on OutdoorRoad and S.MID.
This trend is consistent with the different scaling factors: batch-rebuild depends on the number of points re-clustered, whereas the \mbox{C-ARC} poll floor depends on the number of Clusters (CCs) in $G_C$ and local clusters in $G_P$.
Thus, the relative advantage of incremental polling versus rebuilding is scene- and budget-dependent.

\begin{table}[!t]
    \caption{Median back-to-back poll rate $f_{\max}$ as \mbox{C-ARC} / batch-rebuild (Hz) on Livox Mid-360.
    Higher of each pair in bold.
    Clusters: median per poll at $T_B{=}0.3$~s.}
    \label{tab:fmax}
    \centering
    \footnotesize
    \setlength{\tabcolsep}{3pt}
    \begin{tabular}{@{}l cccc@{}}
        \toprule
        & IndoorOffice1 & OutdoorRoad & OutdoorForest & S.MID~02 \\
        \midrule
        $T_B{=}0.1$~s & \textbf{725}\,/\,144 & \textbf{1464}\,/\,485 & \textbf{893}\,/\,355 & \textbf{465}\,/\,302 \\
        $T_B{=}0.3$~s & \textbf{90}\,/\,37 & 121\,/\,\textbf{137} & \textbf{162}\,/\,87 & 53\,/\,\textbf{83} \\
        Clusters & 762 & 4925 & 7494 & 2009 \\
        \bottomrule
    \end{tabular}
\end{table}

\begin{figure}[!t]
    \centering
    \includegraphics[width=0.95\columnwidth]{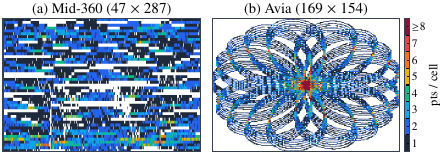}
    \caption{Per-cell occupancy (IndoorOffice1; $T_B{=}0.1$\,s). Empty cells white.}
    \label{fig:occupancy_map}
\end{figure}

\noindent\textbf{Bounding per-cell occupancy.}
The Livox Avia concentrates returns in the FOV centre, so unbounded per-cell occupancy~$k$ inflates neighbourhood insertion (Fig.~\ref{fig:occupancy_map}). 
We implement a per-cell occupancy cap retaining the $c$ most recent returns per grid bucket. 
Overflowing entries are removed from the spatial index prior to neighbourhood search, while remaining in the ring buffer until natural window eviction. 
On IndoorOffice1, $c{=}4$ omits ${\sim}16\%$ of the returns.
Under the same Euclidean protocol as Fig.~\ref{fig:inc_vs_batch}, the uncapped $P_{99}$ poll cycle is $227$\,ms at $T_B{=}0.1$~s and $f_{\mathrm{pub}}{=}20$~Hz (budget $50$\,ms).

Setting $c{=}4$ reduces $P_{99}$ to 41~ms, bringing the system within budget.
For the longer $T_B{=}0.3$~s window, the same cap reduces but does not eliminate the tail, with $P_{99}{=}85$\,ms.
Thus, the occupancy cap is effective at controlling the per-cell insertion cost for short windows but does not by itself guarantee the target budget for larger windows.
These results quantify runtime improvement but not its effect on segmentation quality.

\subsection{Qualitative Analysis}
\label{sec:qualitative_analysis}
\noindent We examine two complementary temporal effects of the sliding window: recovery from ego-motion distortion, and progressive merging of fragmented observations (``healing'').

\begin{figure}[!t]
	\centering
	\subfigure[]{%
		\includegraphics[width=0.325\columnwidth,trim=10 17 10 40,clip]{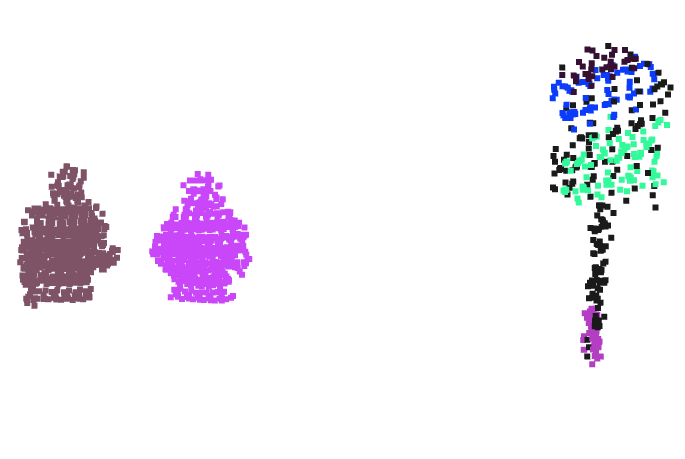}%
	}\hspace{2pt}%
	\subfigure[]{%
		\includegraphics[width=0.325\columnwidth,trim=10 17 10 40,clip]{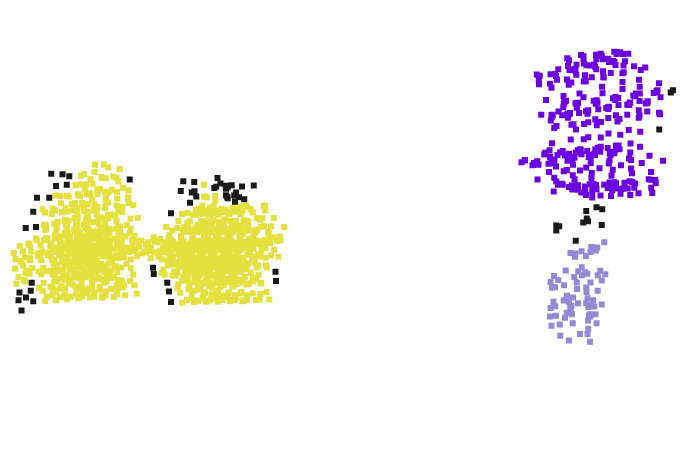}%
	}\hspace{2pt}%
	\subfigure[]{%
		\includegraphics[width=0.325\columnwidth,trim=10 17 10 40,clip]{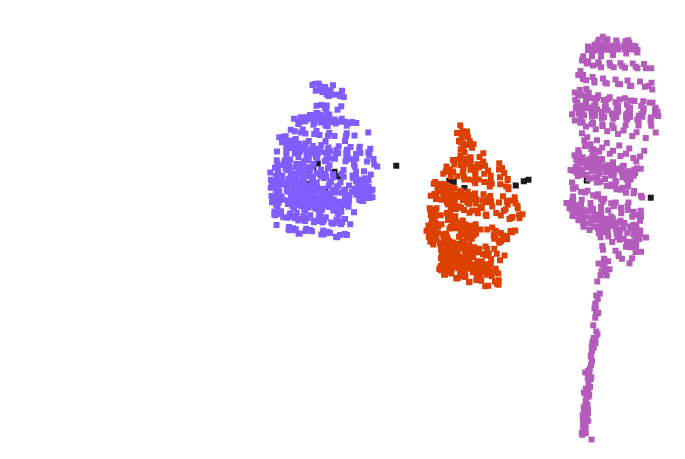}%
	}
	\caption{Recovery from ego-motion-induced distortion. Distinct connected components (a) become incorrectly merged by motion-induced bridge edges (b). As distorted points leave the sliding window, invalid edges are removed and correct separation is restored (c). (Outdoor Forest, Livox Mid-360)}\label{fig:motion_blur}
\end{figure}

Motion-induced bridge edges are particularly relevant for continuous clustering, where incorrect connectivity may persist across insertion cycles.
In frame-based methods, such errors remain fixed for the full duration of $\Delta t_\mathrm{frame}$.
\mbox{C-ARC} bounds their persistence by the eviction schedule of the buffer: once distorted points leave $T_B$, the associated edges are removed and connectivity recovers (Fig.~\ref{fig:motion_blur}).

The complementary filling behaviour is illustrated in Fig.~\ref{fig:healing}: at start-up an object may appear as disconnected fragments~(a); as the buffer accumulates returns, components merge into larger connected structures~(b)--(c).
Connectivity therefore adapts over time: sparse structures densify as observations arrive, while transient motion errors disappear once outdated measurements leave the window.

\subsection{Discussion}
\label{sec:discussion}
\noindent\textbf{Architectural bounds.}
Runtime tracks local occupancy~$k$ and $G_C$ size rather than global point count. 
The Avia's centric pattern limits per-cell occupancy, where depth clipping helps only at short $T_B$ (Sec.~\ref{sec:computational_analysis}).
The Mid-360's omnidirectional pattern avoids these peaks.
Unfiltered ground planes inflate BFS validation; angular linkage~\cite{bogoslavskyi2016FastRangeImagebased} can prevent ground-object merging without dropping returns.

\noindent\textbf{Design choices.}
Scan-pattern dependence is a \emph{runtime} bound, not a modelling assumption: the pipeline avoids prism kinematics and scan-line IDs.
Multiplicity sizing requires only FOV aspect and target load $T_M$, and can seed other grid methods (Fig.~\ref{fig:grid_resolution_comp}, Tab.~\ref{tab:grid_ablation}).
Persistent $G_C$ tracking provides a foundation for temporally stable cluster identities.

\noindent\textbf{Limitations.}
Several scope boundaries frame our evaluation. 
In the experiments, non-Livox sensors validate adaptive grid sizing rather than segmentation quality, whereas spinning-LiDAR evaluations operate per full $360^\circ$ sweep because existing baselines lack support for continuous streaming modes. 
Furthermore, while the per-cell occupancy cap mitigates local density spikes, its downstream trade-off with segmentation quality remains unquantified. 
Lastly, real-time performance conclusions reflect median poll cycles measured during offline dataset replay rather than live, onboard deployment.

\section{CONCLUSION}
\noindent \mbox{C-ARC} maintains a dual-graph over a sliding window and sizes a FOV-aspect bucket range grid from a multiplicity load, without scan-line IDs or prism kinematics.
On a single CPU thread, the Mid-360 meets fixed-rate poll budgets where a per-poll rebuild does not. The Avia's occupancy tail remains the limiter, and a per-cell cap helps only at short $T_B$.
Future work includes native motion undistortion and ground extraction, alternative linkage metrics, and exposing the persistent cluster identities that incremental $G_C$ already maintains.

\bibliographystyle{IEEEtran}
\bibliography{refs}


\end{document}